\documentclass{article}

\usepackage[preprint]{neurips_2022}

\usepackage[utf8]{inputenc} 
\usepackage[T1]{fontenc}    
\usepackage{hyperref}       
\usepackage{url}            
\usepackage{booktabs}       
\usepackage{amsfonts}       
\usepackage{nicefrac}       
\usepackage{microtype}      
\usepackage{xcolor}         
\usepackage{hyphenat}
\usepackage{array}
\usepackage{color-edits}
\usepackage{multirow}
\usepackage{multicol}
\usepackage{tikzlings}
\usepackage[]{graphicx}
\usepackage{natbib}
\setcitestyle{authoryear,open={(},close={)}}

\addauthor{MM}{blue} 
\addauthor{SL}{orange}
\addauthor{nol}{purple} 
\addauthor{YJ}{magenta}
\usepackage[russian,english]{babel}

\newcommand{\lang}{\tikz{\bee[scale=0.15]}}

\newcommand{\vision}{\tikz{\owl[scale=0.15]}}

\title{Measuring Data}

\author{%
  Margaret Mitchell\\
  Hugging Face \\
  Seattle, USA \\
  \texttt{meg@huggingface.co} \\
  \And
  Alexandra Sasha Luccioni \\
  Hugging Face \\
  Montreal, Canada \\
  \texttt{sasha.luccioni@hf.co} \\
  \AND
  Nathan Lambert \\
  Hugging Face \\
  Berkeley, USA  \\
  \And
  Marissa Gerchick  \\
  Hugging Face\thanks{Work done while interning at Hugging Face.} \\
  Palo Alto, USA \\
  \And
  Angelina McMillan-Major \\
  University of Washington \\
  Seattle, USA \\
  \AND
  Ezinwanne Ozoani \\
  Hugging Face \\
  Dublin, Ireland\\
  \And 
  Nazneen Rajani \\
  Hugging Face \\
  Palo Alto, USA \\
  \And 
  Tristan Thrush \\
  Hugging Face \\
  Palo Alto, USA \\ 
    \And
  Yacine Jernite \\
  Hugging Face \\
  Brooklyn, USA  \\
  \And
  Douwe Kiela\\
  Hugging Face \\
  Palo Alto, USA  \\
}

\begin{document}

\maketitle

\begin{abstract}
We identify the task of \textit{measuring data} to quantitatively characterize the composition of machine learning data and datasets. Similar to an object's height, width, and volume, data \textit{measurements} quantify different attributes of data along common dimensions that support comparison. Several lines of research have proposed what we refer to as measurements, with differing terminology; we bring some of this work together, particularly in fields of computer vision and language, and build from it to motivate \textit{measuring data} as a critical component of responsible AI development. Measuring data aids in systematically building and analyzing machine learning (ML) data towards specific goals and gaining better control of what modern ML systems will learn.
We conclude with a discussion of the many avenues of future work, the limitations of data measurements, and how to leverage these measurement approaches in research and practice. \end{abstract}

\section{Introduction} \label{sec:intro}

The size of datasets required for training machine learning (ML) models has quickly grown, as many recent models require amounts of data that are orders of magnitude larger than what was common only a few years ago.  
We are now at a point where many datasets are said to be ``too large to document'' for modern ML systems~\citep{StochasticParrots}. 
Combined with the modern \textit{laissez-faire} approach to dataset development and usage, which is shallow and limited when it is done at all~\citep{LessonsFromTheArchives}, we are currently in the midst of creating, sharing, and training models based on datasets that we know very little about, contributing to behaviour from models that we can neither predict nor trace~\citep{akyurek2022tracing,pruthi2020estimating}.

Measuring data is useful for (1) creating datasets from data sources; (2) documenting existing datasets; and (3) analyzing the outputs of systems as their own kind of data (cf.~\citep{nPMIpaper}), among other uses. By measuring data from data sources, its measurements can be used to determine whether it should (or should not) be included in a dataset. Many measurements can be quickly and automatically calculated, and so help guide data collectors and dataset developers towards creating datasets that meet a given set of requirements or that have well-documented properties.  
For example, by comparing the changes in average sentence length or image size across incremental collection batches, a dataset developer can know whether some data sources may be more preferable to continue sampling from than others for the intended purposes of their dataset. Such real-time data selection can result in better performing models~\citep{lee2021deduplicating}. Applied to datasets that are already built, measurements can be used to document the dataset's characteristics and enable cross-dataset comparison. And applied to the output of machine learning models, such as a language model or the labels generated by an object detection system, measuring data can uncover patterns and biases that the model has learned \citep{nPMIpaper,MeisterCotterell2021}.

 As we discuss below, recent work in fields ranging from philosophy to natural language processing have proposed methods for quantifying data and motivated the importance of this work, yet this research has not yet coalesced around a dedicated task. In the following sections, we detail several proposals for what we refer to as  \textit{data measurements}, and describe their relevance within ML pipelines.

\section{Background and Prior Work}\label{sec:prior-work}

Despite the longevity, ubiquity, and importance of measurement, there is limited consensus on how to define the term, what sorts of things are measurable, or which conditions make measurement possible.  
Different fields have used different terms for measurements somewhat interchangeably, or with different nuances that are not precisely defined~\footnote{Etymologically, while the word \textit{measurement} in English derives from the ancient Greek term $\mu\acute{\epsilon}\tau\rho{o}\nu$ (roughly, \textit{MEH-tron}), \textit{metric} is directly traceable to the use of the suffix -$\iota\kappa o\varsigma$ in Greek: $\mu\epsilon\tau\rho\iota\kappa\acute{o}\varsigma$ (roughly \textit{meh-tree-KOS}), which marks an adjectival form. An informal survey of our peers who speak a variety of Indo-European and Afro-Asiatic languages suggested that different terms for \textit{metric} and \textit{measurement} exist, yet the distinction between them is similarly blurred}. Yet it is important to grapple with its definition to draw together the similar lines of research across fields and to create a common vocabulary and shared understanding around which to organize future interdisciplinary research.

Several fields of research, including those in the physical sciences, metrology, and psychometrics, have offered examples and guidance that can help to further clarify what the task of measuring data can entail. We briefly survey this previous work in this section, and in Section \ref{sec:data-measurements} connect these ideas to recently proposed methods for data quantification. Note that we use the term \textit{dataset} as distinct from \textit{data source}, with the distinction that a dataset is a set of data points created from a data source. Unless otherwise noted, we follow common practice of using the term \textit{data} to refer to both datasets and data sources.

\subsection{A Brief History of Measurement}\label{sec:measurement-history}
\begin{figure}[thb]
    \centering
    \includegraphics[scale=0.075]{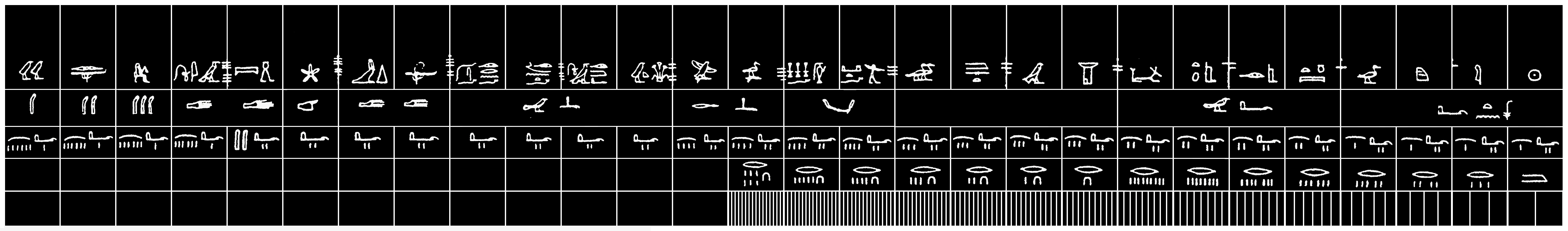}
    \caption{Ancient Egyptian Cubit Rod for measuring length. One of the earliest known objects for standardized measurement. Turin Museum, Wikimedia Commons, CC BY-SA 3.0}
    \label{fig:cubit}
\end{figure}

The idea of \textit{measurement} has existed as far back as we can trace human society. The earliest evidence of measurement is in the form of notches on bones, roughly 33,000 years ago \citep{Vincent22}. Some of the earliest known examples of standardized measurement come from around the 30th century BCE, when the Egyptian cubit marked length based on the distance from the elbow to the tip of the middle finger (see Figure \ref{fig:cubit}).
Some of the first attributes of measurement included length and weight \citep{Michell05,Vincent22}. These are \textit{extensive} attributes of an object, meaning that their value is directly reflective of the structure of the object: the more object there is, the higher the measurement value. Later research on measurement recognized \textit{intensive} attributes, such as temperature, which are not dependent on the amount of the object and whose measurements can be taken \textit{via} extensive attributes. For example, the intensive attribute of density is calculated by dividing the extensive attributes of mass and volume. In the early 20th century, this type of calculation would come to be known as a \textit{derived measurement}, distinct from a \textit{fundamental measurement} such as length.

\begin{figure}
\centering
\includegraphics[scale=0.5]{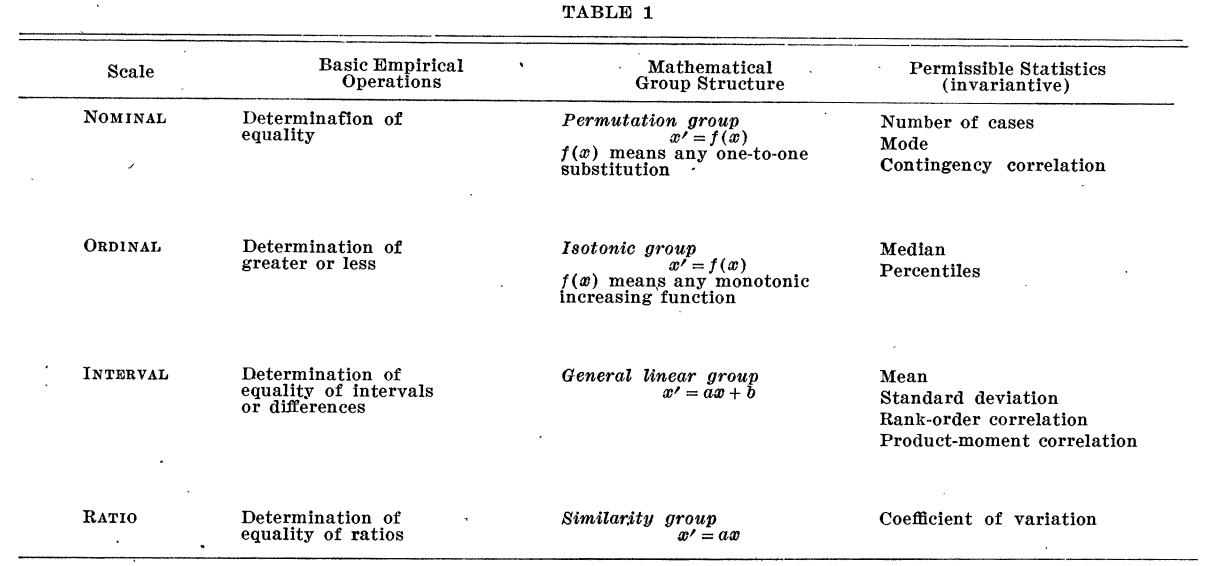}
\caption{S.~S.~Stevens' measurement scales, from the seminal \textit{On the Theory of Scales of Measurement}.
\citep{stevens46}}\label{fig:scales}
\end{figure}

Precise definitions of measurement were put forward throughout the 20th century. While definitions vary, ``the assignment of numbers to items'' is common across many definitions.~\citep{campbell28,stevens46,roberts1985measurement,boslaughwatters08,diez2009history}. Further distinctions include that the assignment of numbers must be systematic (e.g., \cite{roberts1985measurement}), that they facilitate mathematical quantification (e.g., \cite{boslaughwatters08}), that they apply to objects and their properties (e.g., \cite{campbell28}), that they are accessible, proportional, and consistent \cite{Crease2011}, and that they must be gradual, allowing for ``more or less'' comparisons \citep{diez2009history}.
Different scales of measurement have also been defined relatively recently (see Figure \ref{fig:scales}), further deepening our understanding of how measurements can be applied in larger statistical analyses. Precise definitions of measurement paved the way for Measurement Theory, which examines the quantitative structure that abstract concepts, such as comprehension or (problematically) intelligence \citep{harrington1975intelligence}, may take~\citep{michell1997quantitative,hox2005encyclopedia}.

\subsection{Measurement in Machine Learning and Data Science}

The task of measuring data has been at the core of Machine Learning from the early days of the field. In corpus linguistics research adopted into the modern-day ML subfield of natural language processing (NLP), one of the earliest large-scale corpora for English ~\citep{brown1967corpus} was carefully curated and documented with the support of a range of measures, such as by selecting document types to reflect a \textit{prior} base distribution of English written text, or estimating the entropy of the data~\citep{brown1992estimate}. However, as the scale of datasets used in machine learning has drastically increased in recent years, measuring practices have fallen behind as values of efficiency and approaches that prioritize quantity over considerations of other dataset qualities have become more common place~\citep{dataset_politics}.

Addressing this phenomenon, recent work in machine learning has highlighted a critical need for more in-depth probing of the datasets we use for training models, stressing the importance of measures that can enable real-time assessment of data collection to help increase the scientific value and reusability of the data: 

\begin{quote}
``A whole new science of data is needed, with HCI partnership, where sorely needed phenomenological goodness-of-data-metrics need to be developed… Such research is necessary for enabling better incentives for data, as it is hard to improve something we can not measure’’ \citep{DataCascades}
\end{quote}

The idea of ``goodness-of-data'' metrics to measure data resembles recent proposals for quantifying data properties within machine learning research and data science more broadly using different terminology. Examples from research on text and image data include ``quality criteria’’ to analyze, evaluate, and compare the quality of vision and language datasets against one another ~\citep{ferraro2015survey}; ``task-independent metrics’’ to reflect states of the data without the contextual knowledge of the application~\citep{DataQualityAssessment}; ``characteristic metrics’’, which the authors define as ``unsupervised measures’’ to quantitatively describe the properties of a collection of data or a dataset~\citep{LaiEtAl2020}; ``intrinsic metrics’’, to quantify the properties of a dataset \citep{newsroom2018,BommasaniCardie2020}; ``statistical tendencies’’, described as ``present'' in datasets~\citep{MeisterCotterell2021}; ``statistical properties'' to surface issues with dataset contents~\citep{DataAndItsDiscontents}, and ``bias measurements'' to quantify dataset bias~\citep{nPMIpaper}.
Although these different works often do not directly connect their research to one another, they all provide proposals for quantifying datasets and their properties to aid data understanding and comparison. Many proposed quantification methods are also motivated by a shared desire to describe datasets without requiring ``ground-truth'' annotations commonly used in existing training and evaluation pipelines for machine learning models.

The utility of measuring data can be understood against the backdrop of increased ``data-centric AI'' research, with groups in academia and industry stressing the importance of ``systematic methods to evaluate, synthesize, clean and annotate the data used to train and test [AI models]"~\citep{liang2022advances}. This has resulted in proposals for more systematic approaches to data collection~\citep{whang2021data}, data-centric model training~\citep{motamedi2021data}, benchmarking~\citep{eyuboglu2022dcbench}, and debugging~\citep{seal}, as well as hubs that bring together resources and publications in the field~\citep{DCAI}. Recent tools in this vein include CleanLab~\citeyearpar{CleanLab}, DataLab~\citep{xiao2022datalab}  and Snorkel~\citep{ratner2017snorkel}, which provide solutions for analyzing and curating quality ML datasets, but are limited in scope to absent methods for systematic evaluation of the data.

Taken together, this surge in data-centric work makes clear the need for systematic methods of data measurement. Similar thinking extends to tooling provided by and for companies, for example, IBM's tool for data ``quality analysis’’, which uses a series of ``quality scores’’ to enable formalized and systematic data preparation~\citep{jariwala2022data}; Google's ``Know Your Data" tool, created specifically for AI practitioners, has also helped shed light on the contents and characteristics of popular datasets~\citep{KnowYourData}. In terms of AI conferences, the DataPerf workshop was founded specifically for the development of tools and methodologies for dataset analysis \citep{DataPerfNeurIPS2021, DataperfICML2022} and the NeurIPS Datasets \& Benchmarks Track was created in 2021 in order to incentivize dataset creation and analysis~\citep{neuripsdata}, indicating that the field is beginning to recognize the importance of better data quality and its contribution to better performance.  

Prior work has also explored the possible negative consequences of using measurements without proper contextualization or qualification. For example, ~\citet{gururangan2022whose} explore how unexamined definitions of what constitutes data ``quality'' may encode ideologies in ways that further suppress marginalized identities. Such phenomena call into question the possibility of having objective or universally relevant measurements of some implicitly defined data qualities~\citep{dataset_politics}. In order to address these risks when attempting to measure \textit{unobservable theoretical constructs} and particularly social constructs, \citet{jacobs2021measurement} propose a framework for measurement modeling in order to scope the assumptions, purpose, and validity of a specific measurement and avoid misuse or erroneous conclusions.

\section{A Formulation of Measurement for Machine Learning Data}

There is a substantial opportunity to modernize how the data itself is related to advances in ML research by quantifying the composition of the data. We refer to this task as \textit{measuring data}. Intuitively, data measuring can be compared with measurement in the physical sciences. For example, geometry defines what a \textit{distance} is and how it can be used to measure an item's \textit{length}; distance and length measurements can be applied to data in a comparable fashion at the level of the word, sentence, or document, resulting in measurements such as \textit{word length}. Similarly, the measure of the \textit{density} of an object, common in physics, or the \textit{diversity} of specimens in an area, common in biology and ecology, have direct connection to quantifications introduced for machine learning data such as \textit{image density}~\citep{kingma2013auto} and \textit{subset diversity}~\citep{MitchellDiversity}. These newly emerging measurement approaches for machine learning data can leverage the vast history of measurement to further focus the task of \textit{data measurement}.

\subsection{Data Measurements}

An intuition behind \textit{what a data measurement is} is well-explained in Lai et al.~\citeyearpar{LaiEtAl2020}, as something to ``quantitatively describe or summarize the properties of a data collection''. The authors continue:

\begin{quote}
``These metrics generally do not use ground truth labels and only measure the intrinsic characteristics of data. The most prominent example is descriptive statistics that summarizes a data collection by a group of unsupervised measures such as mean or median for central tendency, variance or minimum-maximum for dispersion, skewness for symmetry, and kurtosis for heavy-tailed analysis.'' 
\end{quote}

Following this and the prior work discussed in Section~\ref{sec:prior-work}, we define a \textit{data measurement} as something that:

\begin{itemize}
\item quantifies the magnitude of a characteristic or property of the data
\item is calculated from the data's composition
\item can be composed of, or derived from, units
\end{itemize}

Within the context of data, a \textit{unit} can be a single atomic instance with which a dataset is constructed. For text, this might be the character, word, or sentence. For images, this might be the image or pixel. Following the terminology introduced in Section~\ref{sec:measurement-history}, such units demarcate \textit{extensive} attributes of data for \textit{fundamental} measurements: the more there is, the higher the measurement value.  What can be recognized as \textit{intensive} attributes and \textit{derived} measurements for data have also been proposed. We turn to examples of these measurement types in Section~\ref{sec:data-measurements}, illustrating how these are used in different modalities. 

\section{Examples of Data Measurements} \label{sec:data-measurements}

In the paragraphs below, we describe a set of measurements that have been proposed in different threads of research, contextualized with respect to the specific task of measuring data.

Several approaches are domain- and modality-agnostic, denoted as \textit{General Data Measures}, in order to differentiate from measurements applied specifically in the context of language~\lang~or computer vision~\vision, denoted in \textit{Modality-specific Measures}. Of particular note are measurements that rely on embeddings of the data instances in a metric space, as such representations allow users to more directly leverage the full range of existing measurements in those spaces. Such embeddings may be obtained for example by modeling co-occurrences between data instances through Singular Value Decomposition~\citep{Golub1970SingularVD} or Word2Vec~\citep{mikolov2013efficient}, or by using pre-trained embedding models developed on external datasets. We return to a discussion of the implications of using external sources for dataset measurements in Section \ref{sec:discussion}.

As motivated in Section~\ref{sec:intro}, all of the measurements that we describe can be useful to better understand ML data and datasets, their contents and their characteristics, prior to model training, iteratively during the data collection process, or even after the model is trained, since the output of a model can itself be treated as data and measurements on this data provide quantifications of characteristics that model has learned. The measurements we discuss are both old and new, spanning work on applied statistics within data science as well as research in language and image data analysis. They serve as an example of the kinds of measurements that can be used in the task of measuring data and are summarized in Table~\ref{table:modalities}. We do not proclaim to be extensive -- there are many measurements that we do not describe in the sections below -- but we do aim to illustrate the breadth and multiplicity of data measurements that are relevant to the ML community.

We broadly categorize the measurements as {\sc distance}, {\sc density}, {\sc diversity}, {\sc tendency}, and {\sc association}. We describe them and  summarize their connection to machine learning data in the sections below.

\begin{table}[t]
    \centering
    \small
    \label{table:modalities}
    \caption{%
        Examples of different data measurements proposed in image- and language-based data science and machine learning, alongside analogs in the physical sciences.}
    \renewcommand{\arraystretch}{1.4}
    \begin{tabular}{@{}>{\bfseries}>{\raggedright}p{1.5cm}|>{\raggedright}p{2.1cm}>{\raggedright}p{2.1cm}>{\raggedright}p{2.1cm}>{\raggedright}p{2cm}>{\raggedright\arraybackslash}p{2cm}@{}}
    \toprule
     & \textsc{Distance} & \textsc{Density}   & \textsc{Diversity}      & \textsc{Tendency} & \textsc{Association}\\ \midrule
    Physical Sciences & Length            &  Mass-per-volume & Biodiversity           & Mean, Median, Mode &  Correlation      \\ [.5ex] \midrule
    \multirow{4}{1.5cm}{General Data Measures}   
    & Euclidean Distance & Data Density & Gini Diversity & Burstiness & \\
    & Cosine Similarity & KNN Density & Vendi Score & & \\ 
    & Earth Mover's Distance & & &  & \\
    & Kullback-Leibler Divergence  & & &  &       \\\midrule
    
    \multirow{5}{1.5cm}{Modality-Specific Data Measures}     
    &  Word Mover's Distance (language~\lang) & Information Density (language~\lang) & Text Diversity (language~\lang) & Perplexity (language~\lang) & Pointwise Mutual Information \\
    & Levenshtein Distance (language~\lang)   & {Idea Density}(language~\lang)  & Lexical Diversity (language~\lang) & Fit to Zipf's Law (language~\lang) & \\
    & Inception Distance (vision~\vision) & The Inception score (vision~\vision) &  Image Diversity (vision~\vision)    &    &  \\ 
& & & Subset Diversity (vision~\vision) & & \\\bottomrule
    \end{tabular}
\end{table}

\subsection{Distance}\label{sec:distance}
As something that provides \textit{extensive} values of \textit{fundamental} attributes, distance is one of the most basic measurements. 
Within statistics and data science, \textit{distance} quantifies separation between data instances, variables, distributions, or samples. 
Distance measurements are fundamental to machine learning given that much of ML is about learning representations of data that are easy to handle and manipulate, generally by embedding them in a \textit{metric space} that defines distances between points. 

\textbf{General Data Measures}

\textit{Euclidean Distance:} In its most basic, one-dimensional form, this measurement is familiar across measurement systems for thousands of years and across fields: the length between two points. It can be applied in spaces with two or more dimensions, and so is commonly used when comparing vectors. It can be calculated from the Cartesian coordinates of  points using the Pythagorean theorem, and is symmetric. Applied to data, Euclidean distance can be used to capture everything from how many words are between a re-occurrence of a word, to the divergence between the distribution over words in different parts of a dataset.

\textit{Cosine Similarity:} A measure of similarity between two sequences or vectors, this measurement corresponds to the cosine of the angle between two vectors. This measurement is used in many fields, including NLP, where each word can be assigned a different coordinate and a document can then be represented as the vector of the numbers of occurrences of each word. In this way, cosine similarity can be used to measure of how similar two documents are likely to be in terms of the words they contain. 
    
\textit{Earth Mover's Distance~\citep{rubner2000earth}} represents the minimal cost for transforming one distribution of into another of the same domain. Cost can be defined in different ways depending on the specific applications, for instance either as the Euclidean distance of some measure of the quantity of work necessary to carry out the transformation. Earth Mover's distance has been used in applications ranging from  information retrieval and pattern recognition to calculate the distances between data from different modalities with diverse types of features (e.g.~\citep{peleg1989unified,orlova2016earth}).
    
\textit{Kullback-Leibler Divergence~\citeyearpar{kullback1951information}}: This measure aims to quantify the extent to which one probability distribution is different from another. It is a non-symmetric measure of the difference between the two distribution, i.e. the expected number bits that are necessary to encode points from a given distribution when using another distribution (as opposed to the original distribution directly). Widely used in Bayesian statistics, KL divergence is also useful in applications ranging from magnetic resonance
imaging to analyses of protein and genome structures~\citep{belov2011distributions}.

\textbf{Modality-specific Measures}

Different variations of the general data measurements have also been defined in NLP and computer vision to better reflect the particularities of their modalities and the repercussions that this can have on the way the measurements are calculated.

\textit{Word Mover's Distance}~\citep{kusner2015word} (language~\lang): leverages word embeddings to calculate the minimum cumulative distance from one sequence of words to another. It is particularly useful for measuring the similarity of two documents in a way that general distance measurements do not given that it leverages the geometry of the word space itself. 

\textit{Levenshtein Distance}~\citeyearpar{levenshtein1965} (language~\lang): is a measurement for calculating the distance between strings, defined as the minimum number of single-character edits (i.e. insertions, deletions) that will convert one word into the other. It has been used in language applications ranging from speech recognition to spelling correction given its simplicity and efficiency.

\textit{Inception Distance} (vision~\vision): Fréchet Inception Distance (FID)~\citep{heusel2017gans} and Kernel Inception Distance (KID)~\citep{binkowski2018demystifying}, which leverage an Inception model~\citep{szegedy2015going} trained on ImageNet, were created specifically to assess the quality of images generated by ML models. They do so by calculating the distance between a distribution of generated images with that of a set of real images, considered the ground truth. In that sense, they relate to the tendency measurements that we define in Section~\ref{tendency}, given that they calculate the distances between image distributions, not the images themselves. 

\subsection{Density}\label{sec:density}

\textit{Density} is an qualification of compactness; in the context of datasets, density can be calculated either across a whole dataset or on a subset of it. It indicates how well a data space represents concepts, such as by quantifying how many variations of an exemplar are present. Intuitively, it tells us how well we expect a model to handle a part of the space, or how coherent (possibly implicit) categories of the data are. Density measures have the advantage of being relatively easy to compute across even large datasets, with the disadvantage of being hard to interpret. They can be useful for initial, high-level analyses of datasets and as indicators to guide further fine-grained analyses which can go deeper in terms of the intrinsic characteristics of outliers and representative samples in the dataset (e.g.,~\citep{vig2021summvis,kannan2017outlier}). Relevant measures for density in datasets include:

\textbf{General Data Measures}

\textit{KNN Density}~\citep{loftsgaarden1965nonparametric,knnDensity}: Average similarity between examples, which can be used for determining whether an unlabeled example is an outlier compared to other examples and finding patterns within a dataset. 
Similar to density in the physical sciences, this is a derived measurement that leverages distance; as such, it can also be considered a Distance measurement.
      
\textit{Data Density}~\citep{LaiEtAl2020}: An estimate of the number of samples that fall within a unit of volume in an embedding space. Similarly to density measures in the physical sciences, this measurement utilizes a fundamental measurement -- volume -- to derive a further value. This measurement is particularly useful in approaches involving data visualization, since it can help reflect how many data points are present in a defined area (such as a region of a map). 

\textbf{Modality-specific Measures} 

With the advent of larger and larger datasets, both computer vision and natural language processing have leveraged density measurements to get a bird's eye view of their contents. We describe some of these modality-specific measurements below, given that in practice their definitions can highly depend on their context of application. 
      
\textit{Idea Density}~\citep{IdeaDensity} (language~\lang): Concentration of ideas (or propositions) within a sentence, based on existing research in psycholinguistics which showed that text with a lower density of ideas is easier to understand~\citep{kintsch1973reading}. 
Computing this automatically allows distinguishing between documents that are conceptually simpler (i.e., those written for nonspecialist audiences) versus more technical ones. Different methods may be used to identify an ``idea'' or ``proposition'', which can range from simple word matching within the dataset itself to more complex methods using external models.
      
\textit{Image Density} (vision~\vision): There are many approaches that aim to approximate the probability density functions (PDFs) of images using approaches ranging from modeling the low-level statistics of natural images~\citep{olshausen1996natural} to more complex approaches such as Variational Auto-Encoders (VAEs)~\citep{kingma2013auto}. 
While these approaches can work well for simpler and smaller images (such as the digits from the MNIST dataset~\cite{krizhevsky2009learning}), it remains difficult to model high-resolution images with many objects.
      
\subsection{Diversity}\label{sec:diversity}
Records of measurements quantifying how diverse a sample is are more recent, appearing over the last century in subfields of biology, notably for the purpose of measuring ecological diversity (cf.~\citep{HarrisDiversity,magurran1988ecological}). Diversity is also referred to as heterogeneity in many sciences~\citep{renyi1961measures,nunes2020definition} which have their own approaches to measure it, Hill numbers in ecology~\citep{hill1973diversity}, Hannah–Kay indices in economics~\citep{hannah1977concentration}. In these domains, ``Diversity Indices'' are common, which operate over proportions, entropy (cf.~Shannon Index), and probabilities with respect to randomness (cf.~Simpson Index). The development and usage of diversity measurement within Machine Learning is much more recent (cf.~Vendi Score).

\textbf{General Data Measures}

\textit{Gini Diversity Index}~\citeyearpar{Gini1912}: Known by several different names in fields including economics, sociology, and psychology, this measure has been applied within Machine Learning for classification decision trees to select data splits \citep{raileanu2004theoretical,strobl2007unbiased}, similar to the usage of KL-Divergence discussed in subsection \ref{sec:distance}, although it remains under-utilized in  other approaches and applications.

\textit{Vendi Score}~\citep{friedman2022vendi}: In response to the lack of generic measures of diversity in ML, recent work by Friedman and Deng has proposed a reference-free way to calculate dataset diversity. The Vendi score does this by leveraging the exponential of the Shannon
entropy of the eigenvalues of a similarity matrix calculated based on a user-defined similarity function. As such, it can be applied in fields as spanning from molecular modeling to computer vision and NLP.

\textbf{Modality-specific Measures}

\textit{Lexical Diversity
} (language~\lang): Early research on some of the first ``large'' corpora, such as the British National Corpus, also provided many statistical measurements -- for example, regarding concordances and collocations~\citep{Leech_1992}. This includes 
measurements such as n-gram diversity (or word diversity), which is essentially defined as the number of distinct n-grams or words divided by the total vocabulary~\citep{li2015diversity}. While the recent expansion of dataset size has
witnessed a loss of these measurements, they are useful for many reasons, ranging from detecting
anomalous sequences to determining the maximum input size for models trained on the data.


\textit{Lexical Diversity}~\citep{LaiEtAl2020} (language~\lang): This measure provides a signal regarding how dispersed a cluster of text is, based on a set of embeddings. It can be used as an indicator of dataset homogeneity and to identify subsets or groups within a textual dataset (e.g. data from a given domain or source).
    
\textit{Inception Score}~\citep{salimans2016improved} (vision~\vision): has been used to evaluate the quality of generated images by calculating their diversity based on the 
entropy of the marginal distribution of class
labels as they are predicted by a classifier trained on the ImageNet dataset~\citep{ImageNet}, meaning that it is limited by the classes in ImageNet as well as the quality of its images.
    
\textit{Subset Diversity}~\citep{MitchellDiversity} (vision~\vision): The proportion of human characteristics subject to power differentials in a cluster of data, applied within the context of subset selection of images. This measurement applies specifically to data that represents individuals with respect to characteristics such as gender and race, and requires that these characteristics be known. This can be used to compare the representativity of datasets compared to the populations that they are meant to reflect.
    

\subsection{Tendency Measures}  \label{tendency}

We introduce the term ``tendency'' to categorize summary statistics that quantitatively describe features from a collection of information or a distribution over measurements. 
Tendency values are calculated from distributions over measurements, including count, such as what is captured in descriptive and sufficient statistics. Within machine learning, measurements of central tendency are fundamental, with both objective functions and evaluation scores often utilizing means (averages). Applied to data, tendency measurements can provide information about the general nature of the distributions captured in the data, for example, the length of different instances. 
Common tendency measures include measures of \textit{central tendency} (mean, median, mode) and \textit{dispersion} (standard deviation, variance, the minimum and maximum values of a variable of interest, kurtosis). Other example measures include:

\textbf{General Data Measures}

\textit{Burstiness}~\citep{Burstiness}: Many complex systems can be characterized using intermittent, heterogeneous patterns (``bursts"), properly representing these bursts, their frequency, and amplitude can help predict their future behavior. 
Machine learning datasets that have a temporal aspect (e.g., news articles, product/restaurant reviews, etc.), are the input to such a metric, where modeling the burst patterns can contribute to better understanding the dataset itself.

\textit{Skewness:}~ While assumptions of normal distribution are still commonplace in ML, there are many cases where those assumptions do not hold, especially for datasets gathered `in the wild'. Measuring skewness (i.e. the extent to which the probability distribution of a variable deviates from the normal distribution) can help orient subsequent modeling approaches and indicate whether additional steps are necessary, for instance in terms of class balancing or data normalization~\citep{prati2009data,provost2000machine}. 

\textbf{Modality-specific Measures}

Tendency measures can operate on different units depending on the characteristics of the datasets where they are put into use. For instance, when applied to text datasets, they can operate on levels ranging from characters and words to sentence lengths and frequencies. Applied to image datasets, tendency measures can operate on either the representations of the image pixels or latent representations of the images. 

\textit{Fit to \textit{``Zipf's Law''}}~\citep{MeisterCotterell2021,DMTBlogPost} (language~\lang): A familiar quantity in corpus linguistics and NLP,  measurements based on Zipf's law quantify how closely the quantities of an item in a dataset or data batch match a Zipfian distribution, which is an inverse  rank-frequency distribution. Natural human languages adhere to Zipf's law to various extents, with different coefficients~\citep{gelbukh2001zipf,bentz2016zipf}.  
This coefficient, as well as the distance between the ideal Zipfian distribution and the observed distribution, function as measurement for the naturalness of data \citep{DMTBlogPost}. Recent work has also demonstrated that Zipf's Law arises naturally across domains with underlying, unobserved variables \citep{ZipfLawExtensions}, suggesting it may be extended to datasets in different domains. 
Detecting groups of items that do not correspond to the Zipfian distribution can help identify outliers and undesirable artifacts in a dataset (such as, e.g., HTML tags).

\textit{Perplexity} (language~\lang): Historically, perplexity was used in information theory to reflect how well a probability distribution predicts a sample~\citep{thomas2006elements}. It was subsequently leveraged to evaluate the performance of language models~\citep{brown1992estimate,bengio2000neural,melis2017state} as well as datasets, for tasks such as perplexity sampling~\citep{de2022bertin} and data selection~\citep{toral2015linguistically}. Perplexity can be a useful tool for detecting anomalies in datasets (e.g. sentences that have been incorrectly parsed, encoding errors, etc.), although its limitations have been documented compared to other measures such as Zipf's law~\citep{MeisterCotterell2021}.

\subsection{Association-based Measures}
Measures of association provide quantifications of the relationships between items in a dataset. These were largely developed in the last century, with the most common association measurement. Within machine learning, association measurements provide insight into features that may be redundant with respect to one another, proxy variables, and artefacts of the data that pair concepts together (i.e., whether or not their pairing is reflective of their relationship in the world more generally). One class of association measures that are commonly used throughout work in data science and machine learning are \textit{correlations}, which quantify the degree to which variables are linearly related. 
Other kinds of associations that move beyond the linear relationship include those based on mutual information, such as normalized pointwise mutual information.  

\textbf{General Data Measures}

\textit{Data Correlations}: Common across many fields for years, correlation measures can help quantify how different variables coordinate with one another, i.e., the strength of association and the direction of their relationship.  
The higher the value, the stronger the relationship between them.  
Correlations can be calculated for words within text data, or to signals within vision and audio data, where they can also function as a quantification of similarity.  
There are many choices for measuring correlations: some of the most commonly used measurements are the Pearson correlation coefficient \citep{benesty2009pearson} and the Spearman rank-order correlation coefficient~\citep{ramsey1989critical,kokoska2000crc}, which can help represent inter-dataset relationships and dependencies.

\textit{Pointwise Mutual Information (PMI)}~\citep{nPMIpaper}: PMI provides values for the strength of association between different terms, where a term can be a word in text data, or a label in image data. 
This can be less scalable compared to some of the density- and distribution-based measures described above, as it requires exhaustive pairwise calculations if applied to all terms in a dataset, but is tractable when limited to a set of terms (and their co-occurrences with all other terms). PMI can contribute towards disentangling relationships between terms and detecting patterns and anomalies. 
For example, normalized PMI can be used to identify toxic stereotypes that appear in a dataset as strong associations between identity terms and other items (such as ``woman" and ``smile")~\citep{nPMIpaper}.

\subsection{Other Types of Measurements}

While we have provided an initial categorization of dataset measurements proposed in previous work, there are other existing measures that do not fit into the categories defined above, or that have yet to be applied specifically in machine learning and will be useful to further explore in future research. These include:

\textit{Redundancy Measurements}: Work on redundancy within machine learning datasets has shown the negative impacts of duplicate data on language models~\citep{lee2021deduplicating}, however, quantifying (and removing) duplicated items in a dataset has yet to become a norm in ML dataset creation. Straightforward measurements for this kind of redundancy could include the count of unique instances with duplicates, the total number of duplicates in the dataset, and counts of duplicates by type. Similar to several of the measures described above, entropy can also be used to measure redundancy, such as via the Generalized Entropy Index.

\textit{Readability}~\citep{flesch1979write, halliday1989spoken} (language~\lang): Readability aims to represent how difficult a text is to a reader, and is frequently used in domains such as language didactics~\citep{to2013lexical} and corpus linguistics~\citep{pitler2008revisiting}. It has been proposed as a metric for evaluating natural language generation~\citep{novikova2017we} but has yet to be used for dataset analysis, which could help determine the complexity and linguistic density of texts used for training models. 

\textit{Noise}: Related to the density measures defined above are measures for noise, which can quantify the random variation of density. 
These have been defined for audio and images~\citep{ResearchGateImagePost}. 
They have been defined for text as well (cf.~\citep{VenkataEtAlNoise}), although they tend to require additional models of language. 
Similar proposals to density in text processing include \textit{conceptual cohesion}~\citep{ConceptualCohesion}, \textit{semantic similarity}~\citep{HanEtAlSemanticSimilarity}, and \textit{semantic coherence}~\citep{SemanticCoherence}. 

\textit{Homogeneity and inclusion}: Similarly, many of the authors who have developed diversity measures have also introduced measures of homogeneity~\citep{LaiEtAl2020} and inclusion~\citep{MitchellDiversity}, which are complementary to the diversity measures and can help represent different aspects of (non-)correspondence between the dataset and the populations that it represents, or those that will be impacted by model predictions.

\section{Discussion}
\subsection{On the Limitations of Measurements}

In this work, we have attempted to describe and categorize different approaches for measuring data within the context of machine learning. While all of the measures that we have mentioned can provide crucial information about the composition and contents of datasets, they also come with limitations, both technically and in terms of problematic social biases. Measures and measurements can also be used as a tool of oppression, an issue that must be contended with as we advance work on measuring data that will influence models used throughout society. 

Examining the limitations of measurement through a technical lens, one of the most salient limitations for several of the measurements we describe is that they are usually applied by leveraging an additional source other than the data being measured. For example, FID and KID leverage models trained on ImageNet, whereas perplexity is calculated based on a learned model of natural language. As such, values for these measurements are relative to the external models that are used and do not function to quantify the dataset in-and-of-itself. This means that these measurements are only directly comparable across different datasets when the same models are used -- this is a known issue with perplexity, whose limitations in terms of evaluating language models have been pointed out~\citep{meister2021language}. 
While this can be addressed to some extent by using a standard set of models for computing different measurements, this would not address another severe limitation of these kinds of measures: the issue of bias inherent to the models used for them. For instance, the well-documented biases of the ImageNet categories (see~\cite{crawford2021excavating, luccioni2022bugs}) will influence the results of metrics that rely on these models. This is also true for measures that leverage embedding models (e.g., Word Mover's Distance), since they depend upon the latent representations of concepts from the embeddings, which have been shown to contain problematic biases~\citep{bolukbasi2016man}.

Examining limitations through a social lens, measurements of data, as with measurements of people, have the potential to be influenced by specific values and abstracted away from their original intent. For example, while students' test scores are intended to measure their performance on a specific task, these scores are often used as a proxy to measure a students' individual ability to learn and succeed in later education, and usually do not account for important environmental factors including class size or teacher compensation~\citep{school_tests}.
In fact, assessments are often tied to shifting power dynamics involving implicit discrimination and limiting resources and can be used to justify the usage of algorithms as mechanisms of oppression~\citep{noble2018algorithms}. Measurements provide views on the data, but these views already encode prior beliefs and assumptions about what can be measured and what ought to be measured.
If measures are reported without the right contextualization, they can mislead and provide an undeserved veneer of objectivity to subjective or incomplete evaluations~\citep{impact_factor}.  As such, it is important not to abstract the actual methods that are used to derive a measurement and to directly reference definitions and explain any assumptions used to perform a calculation up front~\citep{jacobs2021measurement}.
This can be done in tools such as datasheets~\citep{gebru2021datasheets} and data statements~\citep{bender2018data}, where measurements can complement other sources of information regarding datasets, such as data sources and metadata.
 
There are also critical lessons that can be learned throughout history on how measurements have been abused by those in power to shortchange the general public. For example, contributors to the \textit{Cahiers de doléances} published at the start of the French Revolution \citeyearpar{cahiersdedoleances} detailed the need for standardized measurements that would create fair pay -- a grievance that led to the creation of the Metric system \cite{NatGeoMetric}.

\subsection{Future Work}
We have discussed measurements with examples largely drawing from the ML subfields of computer vision and natural language processing, where data collection norms are common and notions of measurement are relatively straightforward to connect. However, this only begins to scratch the surface of the space of possible data measurements within machine learning research. 
For instance, in reinforcement learning (RL), where the goal is to enable an agent to maximize a reward signal via trial and error, there are relatively few curated data practices~\citep{sutton2018reinforcement} or standardized quantification strategies. Yet every task comes with a specific reward function that quantifies the agent's performance. These may be possible to conceptualize as types of measurements, which could help with comparing agent actions across different systems. In this example, measuring the data as it returns from the agent has the potential to open new research avenues in how an agent can be improved. Progress in this vein is beginning (cf.~the improved statistical analyses proposed in~\citet{agarwal2021deep}), but substantial opportunity exists to create a comprehensive notion of measurement in line with those developing across other ML sub-fields. 

We have also not addressed measurements outside of datasets comprised of a set of static instances. A dataset comes with collection methodology, norms, judgment protocols, prevailing conditions for measurement, and so is more than a set of instances that can be measured as described here. Datasets also often come with metadata (such as time stamps of collection) that present additional avenues of measurement. Other aspects of datasets require documentation and analyses well beyond quantification, such as annotation contexts \cite{bender2018data,gebru2021datasheets}. 

Our presentation of dataset measurements have also not touched on temporal sequences, a fundamental characteristic of time-series data.  
There are also many other measurements introduced in the physical sciences (e.g., physics-based measures such as energy, brightness, flow) that could potentially be extended to data. Analysing multimodal datasets, such as the recent large dataset LAION~\citep{schuhmann2021laion}, presents unique measurement challenges that we also do not address: Different modalities of a dataset can interact to result in novel harms and biases, as illustrated by~\cite{birhane2021multimodal}. This makes it important to find novel ways of combining existing uni-modal measurements, as well as building upon these to create new measurements that span multiple modalities.

\section{Conclusion}\label{sec:discussion}

We have defined the process of \textit{measuring data} for work in dataset curation and data analysis. The goal of measuring data and datasets is to derive \textit{data measurements}: quantifications of data to understand its composition and the magnitude of different characteristics. The existence of similar lines of research on what we identify as measurement, across fields, suggests the opportunity to foster an interdisciplinary research area on measuring data and datasets, with a shared understanding and common vocabulary. This can aid in understanding what data or a dataset represents, how it compares to other datasets, and how it might be directly improved. We expect that further discussion within research communities will be needed to systematize the definitions and usage of these terms as data measurement develops into an established task.

Previous work makes clear that this task should be approached through perspectives of how to better understand both qualitative and quantitative aspects of data and how to characterize the constructs that the data represents.  This is relevant to processes for developing ML and AI systems, as well as for auditing such systems. Several fields and many researchers have provided helpful insights into different kinds of measurements that can be applied to data; here, we have pulled them together alongside work in fairness and ethics advocating for critical dataset analysis. We believe these lines of work together identify the task of measuring data, motivate its importance, and provide methods for calculating measurements.

Similar to previous work in fairness and ethics in machine learning, we believe concerted efforts in terms of more mindful data and dataset analysis are necessary to pave the way towards higher quality models and tools within the ML community. With the ability to measure data, we gain control over what models learn, and can start to account for problematic artifacts of the relatively uncurated data collection approach used to do, such as overrepresentation of a limited set of viewpoints, or problematic social prejudices encoded as facts. Existing research has proposed several ways to support and empower different kinds of analyses, and many measures and metrics described in previous work are complementary, representing different views of a given dataset. Further work is necessary to directly connect these approaches and to develop the task of measuring data more fully. This would fundamentally advance the rigor in creating datasets, and our ability to understand what they represent.

\section{Acknowledgements}

Thank you to Leon Derczynski and James Vincent for conversations that fundamentally shaped our thinking on this topic.

\bibliography{bibliography}

\end{document}